\documentclass[11pt]{article}
\pdfoutput=1    
    
\usepackage[a4paper]{geometry}
\usepackage{amsmath}
\usepackage{amsfonts}
\usepackage{amssymb}
\usepackage[T1]{fontenc}

\usepackage{booktabs}
\usepackage{natbib,lscape}
\usepackage{pdfpages}
\usepackage{multirow,bigdelim}
\usepackage{algpseudocode} 
\usepackage{graphicx,xcolor,mathrsfs}
\usepackage{setspace}
\usepackage{bbm, bm}
\usepackage{url}
\usepackage[colorlinks,linkcolor=black, citecolor=blue,urlcolor=blue]{hyperref}
\usepackage{varwidth}
\usepackage{multirow}
\usepackage{zref-xr}
\zxrsetup{toltxlabel=true, tozreflabel=false}
\usepackage{cancel}
\usepackage[normalem]{ulem}
\usepackage{subfig, float}
\usepackage{rotating}
\usepackage{array}
\allowdisplaybreaks
\usepackage{fancyvrb}

\usepackage{authblk} 
\usepackage{amsmath, amssymb, graphicx}
\usepackage{natbib} 

\usepackage[utf8]{inputenc} 
\usepackage[T1]{fontenc}    
\usepackage{hyperref}       
\usepackage{url}            
\usepackage{booktabs}       
\usepackage{amsfonts}       
\usepackage{nicefrac}       
\usepackage{microtype}      
\usepackage{xcolor}         

\usepackage{amsmath,amssymb}
\usepackage{amsthm}
\usepackage{amsfonts}
\usepackage{float}
\usepackage{graphicx}
\usepackage{bm}
\usepackage{bbm}
\usepackage{color}
\usepackage{mathtools}
\usepackage{enumitem}
\usepackage{nicefrac}
\usepackage{algorithm}
\usepackage{array}
\usepackage{calc}
\usepackage{wrapfig}
\usepackage{lipsum} 
\usepackage{caption}

\algrenewcommand\algorithmicrequire{\textbf{Input:}}
\algrenewcommand\algorithmicensure{\textbf{Output:}}

\DeclareMathOperator*{\E}{\mathbb{E}}

\newcommand{\wh}{\widehat}
\newcommand{\wt}{\widetilde}

\newcommand{\R}{\mathbb{R}}

\renewcommand{\varepsilon}{\epsilon}
\renewcommand{\tilde}{\wt}
\renewcommand{\hat}{\wh}

\newtheorem{theorem}{Theorem}
\newtheorem{remark}{Remark}

\newcommand{\x}{\mathbf{x}}

\def\eqref#1{equation~\ref{#1}}

\def\1{\bm{1}}

\DeclareMathAlphabet{\mathsfit}{\encodingdefault}{\sfdefault}{m}{sl}
\SetMathAlphabet{\mathsfit}{bold}{\encodingdefault}{\sfdefault}{bx}{n}

\title{F-FOMAML: GNN-Enhanced Meta-Learning for Peak Period Demand Forecasting with Proxy Data}

\title{F-FOMAML: GNN-Enhanced Meta-Learning for Peak Period Demand Forecasting with Proxy Data}

\author[1]{Zexing Xu\footnote{Corresponding author: \href{mailto:zexingx2@illinois.edu}{zexingx2@illinois.edu}}}
\author[2]{Linjun Zhang}
\author[3]{Sitan Yang}
\author[1]{Rasoul Etesami}
\author[1]{Hanghang Tong}
\author[1]{Huan Zhang}
\author[1]{Jiawei Han}

\affil[1]{University of Illinois Urbana-Champaign}
\affil[2]{Rutgers University}
\affil[3]{Amazon}

\date{May 24, 2024}

\begin{document}

\maketitle

\begin{abstract}
Demand prediction is a crucial task for e-commerce and physical retail businesses, especially during high-stake sales events. However, the limited availability of historical data from these peak periods poses a significant challenge for traditional forecasting methods. In this paper, we propose a novel approach that leverages strategically chosen proxy data reflective of potential sales patterns from similar entities during non-peak periods, enriched by features learned from a graph neural networks (GNNs)-based forecasting model, to predict demand during peak events. 
We formulate the demand prediction as a meta-learning problem and develop the Feature-based First-Order Model-Agnostic Meta-Learning (F-FOMAML) algorithm that leverages proxy data from non-peak periods and GNN-generated relational metadata to learn feature-specific layer parameters, thereby adapting to demand forecasts for peak events. Theoretically, we show that by considering domain similarities through task-specific metadata, our model achieves improved generalization, where the excess risk decreases as the number of training tasks increases. Empirical evaluations on large-scale industrial datasets demonstrate the superiority of our approach.
Compared to existing state-of-the-art models, our method demonstrates a notable improvement in demand prediction accuracy, reducing the Mean Absolute Error by 26.24\% on an internal vending machine dataset and by 1.04\% on the publicly accessible JD.com dataset.

\end{abstract}

\vspace{-0.5em}
\section{Introduction}
\vspace{-0.5em}
\label{sec:intro}
Forecasting product demand during high-stake sales events such as Black Friday or Prime Day is a daunting task for both e-commerce giants like Amazon and JD.com and physical retailers. This challenge stems largely from the scarcity of event-specific historical data. Commonly, businesses anchor their strategies on regular sales data, which may not fully capture the distinct consumer behaviors observed during promotional periods. Beyond standard demand prediction, promotional forecasting includes predicting "extreme" events \citep{Laptev2017}. These events, marked by deeper discounts and atypical merchandising strategies, significantly deviate from the typical sales patterns influenced by factors like seasonality or product life cycles. This deviation necessitates a specialized approach to deal-level forecasting, one that thoroughly considers promotion-specific intricacies, from the depth of discounts to deal combinations.

For instance, an online retailer aiming to anticipate the demand spike for a newly launched electronic item during a holiday sale might struggle. They might be unsure how various promotions will influence demand during these events, particularly when previous similar event data is limited or non-existent. To mitigate this, our research effectively uses proxy data from non-peak sales to inform decisions during peak sales events. However, this supplemental data alone is insufficient, given the intricate interrelationships among various products and categories and even across different shopping platforms. We thus introduce a representation learning task for each product, leveraging a cutting-edge Graph Neural Network (GNN) based forecasting model \citep{Yang2023}. This model generates embeddings enriched with graph-enhanced features, encapsulating cross-product information derived from pertinent graph structures. Such structures offer insights into a myriad of dynamics, from relationships between products to patterns of inter-platform shopping behaviors.

Our proposed Feature-based First-Order Model-Agnostic Meta-Learning (F-FOMAML) approach refines the foundational MAML framework~\citep{Finn2017MAML, Nichol2018OnFM}, incorporating task-specific insights~\citep{Yu2022} drawn from GNN-processed data. By training F-FOMAML with this enhanced metadata, the model showcases an unparalleled ability to adapt, consistently surpassing conventional forecasting techniques in accuracy metrics. While we primarily target enhancing e-commerce and brick-and-mortar retail demand prediction, the potential of our GNN-augmented F-FOMAML is vast. Its versatility makes it a candidate for various applications, from fortifying online banking fraud detection systems to optimizing digital advertising click-through rates.

Our main contributions can be summarized as follows:

\begin{enumerate}
\item \textit{Model:} We propose a novel approach to model demand prediction, reframing it as a graph-augmented meta-learning challenge.
\item \textit{Algorithm:} We introduce the GNN-infused F-FOMAML algorithm, which skillfully combines meta-learning and the feature-wise linear modulation (FiLM) layers. This results in a model capable of producing robust predictions, even when historical data is sparse.
\item \textit{Theory:} We provide a theoretical framework that provides insights into how our proposed algorithm reduces predictive risk through the lens of bias-variance trade-off.
\item \textit{Numerical Experiments:} Our numerical experiments address the inherent challenges of forecasting with multi-modal time series data (combining static and dynamic features) while facing data scarcity in both the target domain and source tasks. Empirical tests validate F-FOMAML's proficiency, with the model consistently outshining existing forecasting methods in the prediction MAE values by 26.24\% on the vending machine dataset and by 8.7\% on the JD.com dataset using domain-knowledge constructed features. Furthermore, our model achieves an 1.04\% improvement over the MAE metric against baselines with GNN integrated.
\end{enumerate}

\section{Related Work}
In this section, we discuss prior studies related to sales prediction models, meta-learning's role in time series forecasting, and the significance of proxy data in prediction endeavors. We categorize the related works into four main sub-domains: Prediction with Limited Data, Meta-Learning for Demand Prediction, Few-Shot Meta-Learning Methods, and Graph Neural Networks for Time Series Forecasting.

\paragraph{Prediction with Limited Data}
Previous work in transfer learning has focused on learning from data-rich domains and transferring knowledge to data-sparse regions or underrepresented classes~\citep{transferyield, poverty_mapping,Zhu2020TransferLO}.

In e-commerce, we aim to learn from popular products to improve the performance of new or less popular products. Multi-task learning has also been used to enhance model performance on data-sparse tasks~\citep{rapidresponse, chang2019chimera, fiot2015electricity,pan2009survey}. Conventional transfer learning methods learn transferable latent factors between one source domain and one target domain~\citep{long2013transfer, gong2012geodesic, tzeng2017adversarial,long2015learning}. In our work, we focus on adopting meta-learning techniques to learn from various tasks and then adapt them to unseen tasks in demand prediction.

\paragraph{Meta-Learning for Demand Prediction}
Meta-learning has been applied to various retail and demand prediction tasks, with an emphasis on learning from diverse data sources and adapting to new tasks with limited data. For instance, \citet{retail_demand} employed meta-learning to predict demand in retail settings, demonstrating the effectiveness of meta-learning in capturing complex patterns across diverse scenarios. Similarly, \citet{online_retail} applied meta-learning to online retail data, highlighting the potential for meta-learning in e-commerce applications.

In the time series-related problems, \citeauthor{oreshkin2020n}~\citep{oreshkin2020n} briefly discusses the relation between neural time series prediction and meta-learning~\citep{oreshkin2020n}. \citeauthor{yao2019learning} incorporate the gradient-based meta-learning with a region functionality based memory~\citep{yao2019learning} for spatiotemporal prediction. However, this method relies on the spatial semantic correlations between tasks, which limits its applicability to our problem.

Our work contributes to the problem of learning customer demand for new products with few historical data points. Previous works have suggested comparing the features of new products to existing ones~\citep{ferreira2016analytics, baardman2017leveraging}, or efficient methods for eliciting additional information~\citep{cao2021preference, ma2022constructing}. Our paper assumes that sales have already been observed at limited prices and leverages more information from other related products and environments as proxy data.

\paragraph{Meta-Learning Methods for Few-Shot Learning}
Meta-learning methods for few-shot learning can be broadly categorized into two main approaches: metric-learning-based and optimization-based. Metric-learning-based approaches focus on establishing similarity or dissimilarity between classes, as demonstrated by works such as Prototypical Networks~\citep{snell_prototypical_2017}, Matching Networks~\citep{https://doi.org/10.48550/arxiv.1606.04080}, and Relation Networks~\citep{https://doi.org/10.48550/arxiv.1711.06025}. These methods aim to learn representations that facilitate comparisons between few-shot examples and known classes.
On the other hand, optimization-based approaches aim to learn a good initialization point that can quickly adapt to new tasks with minimal parameter updates. Prominent examples of this category include Model-Agnostic Meta-Learning (MAML)~\citep{Finn2017MAML}, Reptile~\citep{Nichol2018OnFM}, and Meta-SGD~\citep{li2017meta}. These methods have been further extended by advanced techniques such as Task-Specific Adaptation (TSA)~\citep{zhou2020task} and Multi-Modal Model-Agnostic Meta-Learning (MUMOMAML)~\citep{vuorio2019multimodal}. These optimization-based approaches enhance the adaptability and robustness of the learned models across diverse tasks and domains.

Our method falls into the domain of optimization-based approaches.

\paragraph{Graph Neural Networks for Time Series Forecasting}
Deep learning models have been extensively explored for time series forecasting especially those with the Seq2Seq structure \citep{DBLP:conf/nips/SutskeverVL14}, which involves learning an encoder to transform various inputs into fixed-length hidden states for producing forecasts. Recent developments include DeepAR \citep{salinas2020deepar}, TFT \citep{Lim2018} and MQ-Forecasters \citep{wen2017multi, Eisenach2020MQTransform}. However, these methods do not account for cross-observation information, which becomes important in many practical applications. As a result, Graph Neural Networks (GNNs) have rapidly emerged as a promising framework to address this issue by combining temporal processing with graph convolution to augment the learning of individual time series \citep{kipf2016, li2017diffusion,wu2020,shang2021}. A popular family of methods propose graph structure learning for the joint inference of a latent structure through GNN while forecasting \citep{kipf2018,wu2020}. However, they suffer limitations in scaling to large datasets. A scalable approach recently introduced by Yang et al. \citep{Yang2023} uses predefined graphs as data augmentations rather than enabling graph structure learning, which demonstrates not only to scale to graphs over millions of nodes but also shows substantially improving model performance, especially for cold-start problems when data is scarce.   

Our work builds upon these foundations by specifically applying meta-learning and few-shot learning techniques to the demand forecasting problem, with the goal of improving the adaptability and performance of models in this context. To the best of our knowledge, we are the first to study peak period demand prediction with limited records by borrowing relation-aware knowledge from other time periods. We focus on this domain, exploring the application of meta-learning for few-shot prediction and incorporating auxiliary information, such as proxy data from other related tasks, to improve model performance.


\section{Problem Formulation}
\begin{figure*}[!t]
    \centering
    \includegraphics[width=0.9\linewidth]{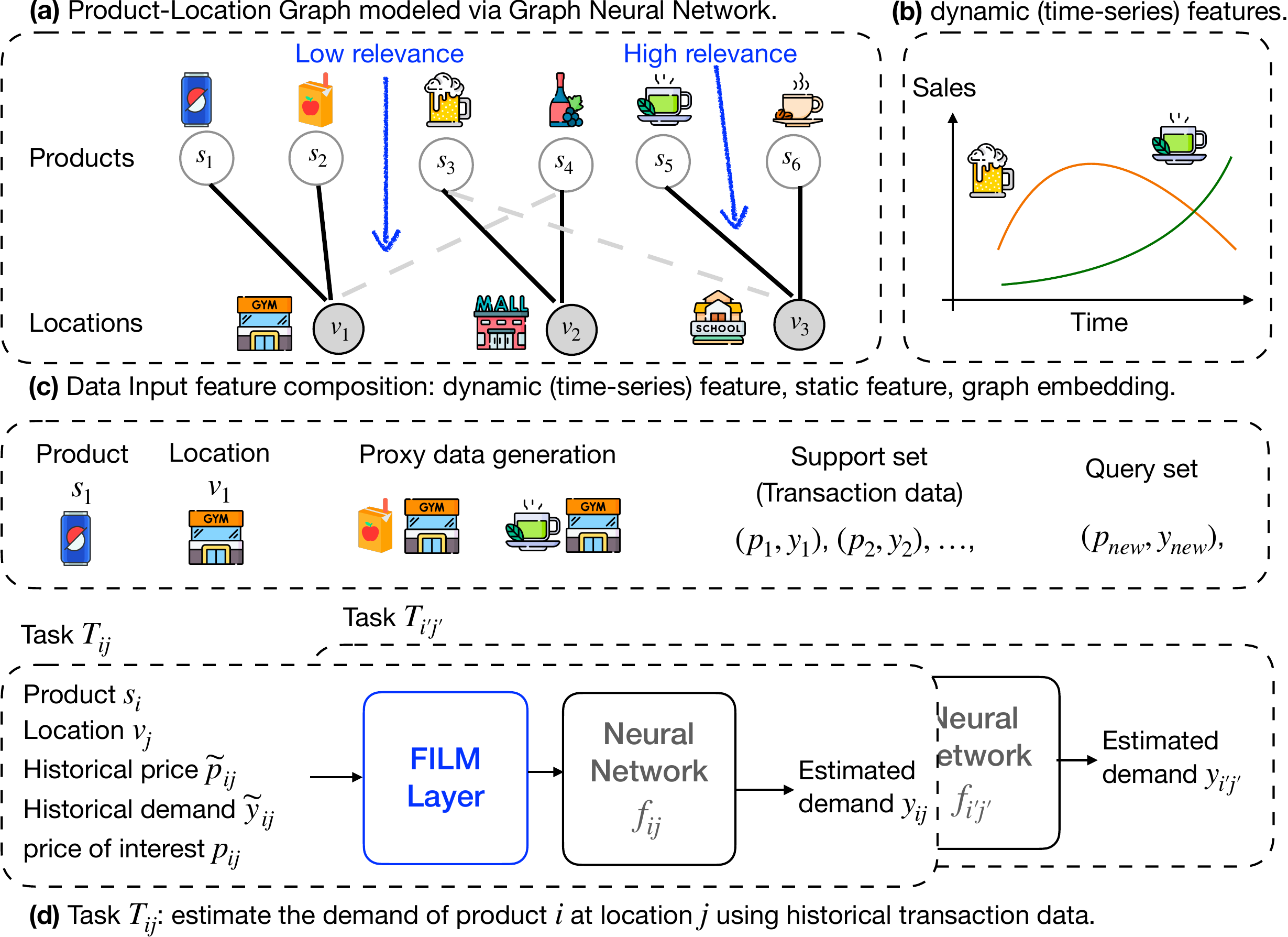}
    \caption{Pipeline of the GNN-enhanced F-FOMAML for demand forecasting. }
    \label{fig:film-pipeline}
    \vspace{-2em}
\end{figure*}

During peak periods, promoted products often have limited historical sales (e.g., less popular items) or are new items without historical transaction data. Consequently, the demand forecasting tasks for product-location pairs during these periods are new and unseen compared to regular products and periods (e.g., paper towels). To address this challenge, we frame our research problem within a generic setting, focusing on a few-shot meta-learning paradigm, specifically targeting demand forecasting. Throughout our discussion, we use JD.com's transactional data as the primary example to illustrate our approach.
\subsection{Task Definition}
Demand forecasting aims to predict the future demand for a product in a specific environment based on observed features. Each forecasting task is associated with a product and its environment.

Formally, let $\mathcal{P}({\mathcal T})$ denote a distribution over tasks ${\mathcal T}_{ij}$, each corresponding to product $i$ in environment $j$. For a set $[n]=\{1,\ldots,n\}$ of $n$ products with product $i$ present in $t_i$ environments, we have a total of $\sum_{i=1}^n t_i$ tasks. Each task dataset is symbolized as a pair $(\mathbf{x}_{ij}, y_{ij})$, where $\mathbf{x}_{ij}$ is the feature vector and $y_{ij}$ signifies the associated demand.


To provide a concrete example, consider a scenario where we have $n=10$ products, each available in $t_i=5$ locations. Therefore, we have a total of 50 tasks in our meta-training set. The dataset corresponding to each task is represented as a demand-feature pair $(\mathbf{x}_{ij}, y_{ij})$, where $\mathbf{x}_{ij} \in \mathbb{R}^m$ is a feature vector and $y_{ij} \in \mathbb{R}$ is the associated demand.

Our goal is to train a model, denoted by $f: \mathbb{R}^m \rightarrow \mathbb{R}^+$, capable of mapping $m$-dimensional observations $\mathbf{x}$ to outputs $y$ across a large or possibly infinite number of tasks. We employ the First-Order Model Agnostic Meta-Learning (FOMAML) algorithm for this purpose. For a given product characterized by a feature vector $s_i, \forall i \in [n]$ and an environment (e.g., location) characterized by a feature vector $v_j, \forall j \in [t_i]$, we consider a single historical price and demand observation $(\Tilde{p}_{ij}, \Tilde{y}_{ij})$.

Given a price of interest $p_{ij}$, we assume our task as the following demand function:
\begin{equation}\label{eq:Demand}
y_{ij} = f_{ij}(\mathbf{x}_{ij})+\epsilon_{ij},
\end{equation}
where $\mathbf{x}_{ij}$ is the feature tuple $(s_i,v_j,\Tilde{p}_{ij},\Tilde{y}_{ij}, p_{ij})$ and $y_{ij}$ is the corresponding demand $y_{ij}$. Here, $f_{ij}$ is a flexible function (e.g., linear regression, MLP, etc.) and each task is associated with a unique model parameter $\beta_{ij}\in{\mathbb R}^m$. We assume that the noise $\epsilon_{ij}$ follows a centered sub-Gaussian distribution with parameter $\sigma_i^2$. Furthermore, without loss of generality, we assume that $\mathcal{P}_{X}$ is an isotropic-centered sub-Gaussian distribution, i.e., $\mathbb{E}(\mathbf{x}_{ij}\mathbf{x}_{ij}^\top) = \mathbb{I}_d$. Exploiting some structural similarities in $\mathcal{P}\left({\mathcal{T}}\right)$, the goal is to train a model for a new task ${\mathcal{T}}^{\rm new}$, coming from $\mathcal{P}\left({\mathcal{T}}\right)$, from a small amount of {\em training dataset} ${\mathcal{D}}={\big(\mathbf{x}^{\text{new}}_{ij},y^{\text{new}}_{ij}\big)}$.

\begin{remark}
Incorporating features allows us to capture an additional form of shared structure. However, despite accounting for observed product features, the demand functions of two products can exhibit distinct behaviors. For instance, even for Diet Coke, price sensitivities may vary significantly on different vending machines due to factors such as customer demographics or preferences that are challenging to capture as explicit features. To account for these product-location-specific nuances, we introduce the flexibility for the demand function's \textit{coefficients} (e.g., price elasticity) to differ.
\end{remark}

In the First-Order MAML (FOMAML) approach, the model parameters for each task in the meta-training dataset are computed after a single gradient update. Specifically, for each task $\mathcal{T}_{ij}$, the task-specific model parameters, denoted $\beta_{ij}^{\prime}$, are updated as follows:
\begin{equation}\label{eq:task}
\beta_{ij}^{\prime} \leftarrow \beta^* - \lambda \nabla_{\beta^*} \mathcal{L}_{\mathcal{T}_{ij}}(\beta_{ij}),
\end{equation}
where $\lambda$ is the learning rate, $\beta^*$ is the global model parameter shared across tasks, and $\mathcal{L}_{\mathcal{T}_{ij}}(\beta_{ij})$ is the task-specific loss, such as the mean squared error:
\begin{equation}
\mathcal{L}_{\mathcal{T}_{ij}}(\beta_{ij}) = \frac{1}{2} \left( y_{ij} - f_{ij}(\mathbf{x}_{ij}) \right)^2.
\end{equation}

After updating the task-specific parameters, a meta-update is performed on the shared global parameter $\beta^*$ using the performance of the updated $\beta_{ij}^\prime$ on their corresponding tasks. This meta-update is given by the following:
\begin{equation}\label{eq:meta}
\beta^* \leftarrow \beta^* - \eta \sum_{i=1}^{n}\sum_{j=1}^{t_i} \nabla_{\beta^*} \mathcal{L}_{\mathcal{T}_{ij}}(\beta_{ij}^{\prime}),
\end{equation}
where $\eta$ is the meta-learning rate. The objective of this meta-learning process is to optimize the shared global parameter $\beta^*$ such that, after a few updates on each task, the task-specific parameters $\beta_{ij}$ yield improved performance on their corresponding tasks. Once the meta-learning process is complete, the model parameters of a newly arriving task can be estimated using the learned meta-parameters $\beta^*$. These task-specific parameters $\beta_{ij}^{\prime}$ can then be fine-tuned on the new task using the available data, yielding improved performance and adaptability to new tasks.

By incorporating the FOMAML algorithm into our meta-learning framework, we aim to construct an efficient model for sales prediction that can adapt to new tasks with limited historical sales data.

\section{Methodology} \label{sec:method}
We illustrate the pipeline of our algorithm in Figure~\ref{fig:film-pipeline}. Imagine there are 3 locations offering 6 drink types with transaction data, capturing their historical sales. First, a graph neural network (GNN), $G$, is formed using both static features like machine locations and dynamic features from past sales time series. To predict the demand for Coke at the gym with a discount, relevant nodes and edges from $G$ are extracted. This subset, denoted as 
$G_\mathcal{T}$, undergoes training using MAML's inner loop, yielding initial task-specific parameters. These parameters are further refined through the FiLM transformer, considering proxy data that might suggest a promotion for Coke. The shared meta-parameters are updated in MAML's outer loop based on the specific task losses. Once this cycle is completed across all tasks, the model is evaluated on fresh data to project the demand.

Our proposed methodology for e-commerce demand prediction encompasses three pivotal components: proxy data selection, GNN-enhenced representation learning, and the F-FOMAML algorithm design. To cater to the multi-faceted nature of e-commerce products and their varied demand across different locations or customer segments, we weave task-adaptive estimators into the meta-learning framework. Further, we employ GNN and the FiLM layer to utilize and encode proxy data into hidden representations, thus enabling the modulation of learner parameters for enhanced adaptation to the specific characteristics of products and customer segments.
\subsection{Proxy Data Selection}
\label{sec:proxydata}
The proxy data, vital for tasks with limited historical sales data, is judiciously selected. The ideal proxy data simulates the potential sales behavior of the focal product, informed by sales trends of similar products or those in related categories.

For e-commerce scenarios, task similarity might arise from: 1) \textit{Historical Transactions}: Edges represent products often purchased together. 2) \textit{User Behavioral Patterns}: Edges might indicate similar purchase behaviors or browsing patterns of users. 3) \textit{Product Similarities}: Linking products of the same category or with similar attributes. 4) \textit{Domain Knowledge}: Connections deriving from expert insights into customer behaviors, seasonal trends, or market dynamics.

Traditional methods use clustering techniques and measure distances with metrics like Euclidean and cosine similarity to quantify task similarity. Our approach, however, leverages a GNN-based method for selecting relevant tasks as proxy data.
For a given task, we denote its proxy data as $Z_{ij}$, representing the most relevant data identified through our GNN-based approach. This ensures the proxy data accurately reflects the target task, enhancing demand forecasting accuracy.
\paragraph{Graph Construction for Proxy Data}
A tailored graph for our GNN encapsulates relationships among tasks. In determining proxy data for e-commerce settings, we choose tasks from support set $\mathcal{T}$ resembling our target task, $\mathcal{T}_{\text{new}}$, guided by 
$\text{correlation}(\mathcal{T}, \mathcal{T}_{new}) > \delta,$
where $\delta$ is a threshold indicating task similarity, and the function $\text{correlation}(\mathcal{T}, \mathcal{T}_{new})$ captures the similarity between $\mathcal{T}_{new}$ and $\mathcal{T}$ through different methods such as the ones described above.
\subsection{GNN-enhanced Representation Learning} 
\label{sec:gnn-construct}
Here, we describe how to obtain the graph-enhanced features for each product. In a nutshell, we set up a time series forecasting task and utilize a GNN-based demand forecasting model to predict future sales given each product's historical information as well as cross-product relationships defined by a predefined graph. We then extract the hidden encoded context from the trained model to produce the product embeddings as features.  
\vspace{-1em}
\paragraph{Input Product Features}
E-commerce platforms host a plethora of products, each with unique characteristics and consumer interactions. In this case, We construct the graph using product-specific attributes such as brands (i.e., we connect all products with the same brand). The input features for node $N_i$ (representing product $i$) are: 1) Static features $S_i$,  like the product category, brand, and manufacturing details. 2) Dynamic features $D_i$, encompassing time-evolving aspects like recent sales and price changes.
\paragraph{Product Embedding Generation via Forecasting}
A crucial aspect is to generate meaningful product embeddings that can capture the multifaceted nature of e-commerce products. To facilitate this, we set up a demand forecasting task as:
\begin{equation}
\label{eqn:fctask}
\widehat{Y}_{t+1} = f\left(Y_{t-C:t}, D_{t-C:t}, S\right),
\end{equation}
where $f$ represents the forecasting model. At time $t$, target $Y_{t+1}\in\mathbb{R}^{N\times 1}$ are the future one-day sales, $D_{t-C:t} \in\mathbb{R}^{N\times d}$ are $d$ dynamic features with the history length of $C$ days, and $S\in \mathbb{R}^{N\times m}$ are $m$ static features for all $N$ products. We adopt the GNN-based forecasting model introduced in \citep{Yang2023} and use the brand information to craft the predefined graph. The GNN is utilized both for forecasting and for generating the embedding of tasks. After training convergence, we extract the embedding for each product, which serves as compact representations of product dynamics.
\paragraph{Edge Relationship Determination}
Let $E(\mathcal{T}_i, \mathcal{T}_j)$ denote the edge between tasks $\mathcal{T}_i$ and $\mathcal{T}_j$.
The edge relationships between the two entities are inferred using:
\vspace{-0.6em}
\begin{equation}
 E(\mathcal{T}_i, \mathcal{T}_j) = 
\begin{cases} 
1 & \text{if } \text{dist}(emb(\mathcal{T}_i), emb(\mathcal{T}_j)) < \theta \text{ or } h_{\mathcal{T}_i} = h_{\mathcal{T}_j} \\
0 & \text{otherwise,}
\end{cases}
\end{equation}
where $emb(\mathcal{T}_i)$ stands for the embedding of task $\mathcal{T}_i$, $\text{dist}(\cdot,\cdot)$ denotes a function measuring the distance between two embeddings,  $h$ denotes the task (i.e., product) hierarchy or taxonomy, and  $\theta$ is a pre-determined threshold to determine closeness. We will create an edge between tasks $\mathcal{T}_i$ and $\mathcal{T}_j$ if either their corresponding embeddings are close to each other or they belong to the same category.

\subsection{F-FOMAML Algorithm Description}
We present the GNN-Integrated Feature-based First-Order MAML (F-FOMAML) for Demand Forecasting Algorithm~\ref{alg:f_FOMAML}, which incorporates transactional data with static and dynamic features, and proxy data to forecast demand in peak periods. This variant of the MAML algorithm delineates the learning process into several stages: meta-learner, base learners, FiLM Layer, and fine-tuning.

\begin{algorithm}[!t]
\caption{GNN-Integrated Feature-based First-Order MAML (F-FOMAML) for Demand Forecasting} \label{alg:f_FOMAML}
\begin{algorithmic}[1]
\Require{$\mathcal{D}$=\{Static features $S=\{(s_i,v_j)\}_{i,j}$, Dynamic features $D=\{\Tilde{y}_{ij}\}_{i,j}$; Proxy data $Z$; pre-constructed graph $G=(N, E)$; learning rates $\eta,\lambda$.}
\State Learn the GNN with nodes $N$ and edges $E$ using $S$ and $D$.
\State Initialize global meta-parameters $\beta^*$.
\For{each task $\mathcal{T}_{ij}$ in $\mathcal{D}$}
    \State Extract relevant nodes $N_{\mathcal{T}_{ij}} \subset N$ and edges $E_{\mathcal{T}_{ij}} \subset E$ from $G$ for product~$i$ in environment~$j$
    \State Represent task $\mathcal{T}_{ij}$ using a subgraph $G_{\mathcal{T}_{ij}} = (N_{\mathcal{T}_{ij}}, E_{\mathcal{T}_{ij}})$.
    \State Initialize task-specific parameters $\beta_{ij}$ from $\beta^*$.
    \State Compute task-specific loss $\mathcal{L}_{\mathcal{T}_{ij}}(\beta_{ij})$:
    $
    \mathcal{L}_{\mathcal{T}_{ij}}(\beta_{ij}) = \frac{1}{2} \left( y_{ij} - f_{ij}(\mathbf{x}_{ij}; \beta_{ij}) \right)^2.
    $
    \State Update task-specific parameters $\beta_{ij}^{\prime}$ using Equation~\ref{eq:task}:
    $
    \beta_{ij}^{\prime} \leftarrow \beta^* - \lambda \nabla_{\beta^*} \mathcal{L}_{\mathcal{T}_{ij}}(\beta_{ij}).
    $
\State Apply FiLM transformer using proxy data $Z_{ij}$ to modulate the input features $\x_{ij}$ for task $\mathcal{T}_{ij}$, enhancing the task-specific parameters $\beta_{ij}^{\prime}$ adaptation:
$\text{FiLM}(\x_{ij}; Z_{ij}) = \beta(Z_{ij}) \odot \x_{ij} + \gamma(Z_{ij})$

\EndFor
\State Perform meta-update on $\beta^*$ using:
$
\beta^* \leftarrow \beta^* - \eta \sum_{i=1}^{n}\sum_{j=1}^{t_i} \nabla_{\beta^*} \mathcal{L}_{\mathcal{T}_{ij}}(\beta_{ij}^{\prime}).
$
\State Evaluate the model on testing data to get forecasted demand $\hat{y}$.
\State \Return Forecasted demand $\hat{y}$ for products. 
\end{algorithmic}
\end{algorithm}

\paragraph{FiLM Layer.}
The feature-wise linear modulation (FiLM) layer~\citep{perez_film_2017} is a critical component in tailoring the learner parameters based on the proxy data features. This layer applies an affine transformation, feature-wise, to its input, modulating the hidden vector outputs of the meta-model using the proxy data $Z_{ij}$ as task encodings. The construction and purpose of the proxy data $Z_{ij}$ is elaborated in section \ref{sec:proxydata}. The FiLM layer facilitates a more refined adaptation to the distinctive traits of the product and vending machine location by exploiting the relationship between the product-specific and machine-specific price-sensitivity estimators encapsulated in the proxy data.

The FiLM layer's mechanism can be formally described as
\begin{equation}
\text{FiLM}(\x_{ij}; Z_{ij}) = \beta(Z_{ij}) \odot \x_{ij} + \gamma(Z_{ij}),
\end{equation}
where $\x_{ij}$ represents the input feature representation, $\beta(Z_{ij})$ (abv. $\beta_{ij}$) and $\gamma(Z_{ij})$ signify the scaling and shifting factors learned from the proxy data $Z_{ij}$, and $\odot$ symbolizes element-wise multiplication. The functions $\beta(Z_{ij})$ and $\gamma(Z_{ij})$ are learned during the training phase to cater to the specific task at hand. By applying this transformation to the task-specific model parameters $\beta_{ij}^{\prime}$, the model captures complex feature interactions and becomes better equipped to adapt to the specific characteristics of each unique product-environment pair.
\paragraph{Meta-Learner.}
The core of the meta-learning approach is the meta-learner, an overarching model that helps in initializing and updating the meta-parameters, \(\beta^*\). These parameters serve as a shared knowledge base that aids in swift adaptation across a myriad of tasks. The meta-learner initializes the global meta-parameters $\beta^*$ and, after task-specific adaptations are performed, updates $\beta^*$ using the aggregated first-order gradients from each task. This process ensures that the meta-parameters incorporate insights from various tasks, enabling rapid adaptation to new tasks and reducing the cold-start problem in the e-commerce domain.
\paragraph{Base Learners.}
The base learners are models tailored to specific tasks, such as predicting the demand for a new product launch or forecasting sales during a flash sale. Each base learner operates by extracting relevant nodes and edges to form a subgraph for each task, initializing task-specific parameters $\beta_{ij}$ from the global meta-parameters $\beta^*$, computing the task-specific loss, and updating the task-specific parameters $\beta_{ij}'$ using first-order gradient descent. Additionally, the FiLM transformer applies proxy data to modulate input features, enhancing the adaptation of the task-specific parameters.

\paragraph{Fine-Tuning.}
In our approach, fine-tuning involves a final meta-update on $\beta^*$ after task-specific updates, evaluating the model on testing data to forecast demand $\hat{y}$ for products and returning the forecasted demand $\hat{y}$ as the final prediction output. This process leverages the FiLM transformer and proxy data to ensure that the models are not just generic but tailored to capture the heterogeneity in tasks.

The strength of this method mainly lies in its ability to utilize shared structures across tasks while also adapting swiftly to unique task characteristics using the FiLM transformer and proxy data.
\section{Theoretical analysis}\label{sec:theory}
In this section, we provide a theoretical model to illustrate the benefit of our proposed method and shed light on why proxy data improves the few-shot prediction.
\paragraph{Data generative model.}
Suppose we have a set of tasks $\mathcal T$. For each task $t\in\mathcal T$, we observe training samples $\{(x_k^{(t)},y_k^{(t)})\}_{k=1}^{n_t}$, where $n_t$ is the sample size for the task $t$.  
 In addition, for each task, we observe a task-specific feature $v_t\in\R^r$. For the simplicity of presentation, we assume $v_t\in [0,1]^r$.
 We denote the set of training domains by $\mathcal D^{tr}$ and assume there are ${T}=|\mathcal{T}|$ training tasks. Following equation (\ref{eq:Demand}), we assume that for each task $t$, the outcome prediction function $g$ takes the form of $
y=g_t(x)+\epsilon:= h_t(f(x))+\epsilon$,
where $h_t$ is the base-learner that depends on individual task $t$, $f$ is the meta-learner, and $\epsilon$ is a noise term which is assumed to be sub-Gaussian with mean 0 and variance $\sigma^2$.

Following Section~\ref{sec:method}, for each task $t\in\mathcal T$, we construct the proxy data $Z_{t}$ by including all similar tasks $t'$ such that $\|v_{t'}-v_t\|\le h$ for some threshold parameter $h>0$. Then similarly, for the test task $\tilde t$, the outcome prediction function $\hat g_{\tilde t}$ is computed as $
\hat g_{\tilde t}(x)=\hat h_{\tilde t}(f(x)),
$
where $\hat h_{\tilde t}(\cdot)=\frac{\sum_{t\in\mathcal T} w(\tilde t, t) \hat h_i(\cdot)}{\sum_{t\in\mathcal T} w(\tilde t, t)}$, with the weight $w(\tilde t,t)=\1\{\|v_{\tilde t}-v_t\|\le h\}$. In the case where the denominator is $0$, we define $\hat h_{\tilde t}=0$.

\paragraph{Theoretical results.}
To facilitate the theoretical analysis, we first assume that the distance between task-specific features indeed captures the similarity of tasks: there exists a  universal constant $C$, such that $
\| h_{t_1}- h_{t_2}\|_\infty\le C\cdot \|v_{t_1}-v_{t_2}\|.
$

In addition, we assume that for each training domain $t$, $\hat h_t$ is well learned such that $\E\big[\big(\hat h_t(f(x))-h_t(f(x))\big)^2\big]=O(\frac{C(\mathcal H)}{n_t})$, where $C(\mathcal H)$ is the Rademacher complexity of the function class $\mathcal H$.  We further assume $v_t$ has a positive density over $[0,1]^r$. Then, we have: 

\begin{theorem}\label{thm:excess-risk}
Consider the data generative model, the algorithm $\hat g_{\tilde t}$, and the assumptions above. Suppose we have $n_d\gtrsim n$ for all $t\in\mathcal D^{tr}$. Define the excess risk for the test domain $\tilde t$ by $R(\hat g_{\tilde t})=\E_{(x,y)\sim \tilde t} [l(\hat g_{\tilde t}(x;\mathcal D^{tr}, A),y)]-\E_{(x,y)\sim \tilde t} [l(g_{\tilde t}(x;\mathcal D^{tr}, A),y)]$.
If the loss function $l$ is Lipschitz with respect to the first argument, then the excess risk satisfies
\vspace{-0.6em}
\begin{equation}\label{eq:1}
    R(\hat g_{\tilde t})\lesssim h+\sqrt\frac{C(\mathcal H)/n}{\max\{1,nh^r\}}.
    \vspace{-1em}
\end{equation}
In particular, if $h$ is properly chosen such that $h\asymp (\frac{C(\mathcal H)/n}{T})^{\frac{1}{r+2}}$, then
$
R(\hat g_{\tilde t})\lesssim \left(\frac{C(\mathcal H)/n}{T}\right)^{\frac{1}{r+2}}.
$
\end{theorem}
Theorem~\ref{thm:excess-risk} suggests that the superiority of our algorithm comes from a better bias-variance trade-off. More concretely, the threshold $h$ tunes the trade-off for the excess risk. On the one hand, when we do not use relational data at all (corresponding to the case where $h= 0$), the first term in \eqref{eq:1}, bias, is negligible, while the second term, variance, becomes dominant as the data is limited. As the excess risk of a single task is of order $(C(\mathcal H)/n)^{1/2}$, our result implies that when $T$ is sufficiently large such that $T> \big(\frac{n}{C(\mathcal H)}\big)^{r/2}$, our proposed method will overcome the potential bias by incorporating similar (but still different) tasks and better than learning with only one single-source task. On the other hand, if we simply use the standard ERM to pool all the data together  (corresponding to $h=\infty$), although the variance becomes small, the bias would dominate in this case. The second part of the theorem suggests that one can efficiently incorporate the proxy data with a carefully chosen threshold. The proof of Theorem~\ref{thm:excess-risk} is deferred to Appendix~\ref{sec:proof}. 


The theoretical perspective we discussed is particularly pertinent to the context of demand prediction. Given the limited data available from high-stakes sales events, relying solely on this data for prediction (akin to $h=0$) can result in outcomes with substantial variance. Conversely, utilizing the entire historical dataset (corresponding to $h=\infty$) can introduce significant bias, given the marked differences between regular and high-stakes sales events. Our method harnesses GNNs to understand the relationships within historical data. This approach strikes an optimal balance in the bias-variance trade-off, leading to improved prediction accuracy.

\section{Experiment}\label{sec:exp}
In this section, we conduct extensive experiments to evaluate the efficacy of our proposed F-FOMAML for peak-period demand prediction, focusing on two key research questions: 
\begin{itemize}
    \item How does F-FOMAML's prediction performance compare to various baselines? 
    \item To what extent do the components we introduce, such as the proxy data selection method (GNN versus MQCNN), impact the model's predictive capabilities?
\end{itemize}
By addressing these questions, we provide a comprehensive evaluation of F-FOMAML, highlighting its performance relative to existing approaches and analyzing the contributions of individual components to the model's overall predictive power.

\subsection{Experimental Setups}\label{sec:exp_setup}

In this section, we detail the experimental setups, focusing on two real-world datasets and the evaluation criteria for our method's performance. 

For brevity, the main text covers the data description, experimental setup, and results for the JD.com dataset, while the details for the vending machine dataset are provided in the Appendix~\ref{sec:add_exp}.

\paragraph{Datasets.} We validate our methodology using transactional records from JD.com~\footnote{Dataset available at: \url{https://connect.informs.org/msom/events/datadriven2020}}, which include both static and dynamic features related to products (SKUs) and order details for March 2018. 

The goal is to predict the demand at the promotional price given the demand at the regular price. We use the category information (3 categories in total) for product features, and region (63 regions) information for location features. We use the last 15 days as testing, and the second to last 15 days as training. The detailed data description and dataset construction are deferred to the Appendix~\ref{sec:dataset}.

\begin{table}\label{sec:result_analysis}
    \centering
     \caption{Experiment results on real-world JD.com E-Commerce sales data with proxy data features generated from different methods. This includes regression techniques, ensemble strategies, neural-network-based methods, and transfer methods for a comprehensive benchmark.}
    \label{tab:jd_result}
\begin{tabular}{@{}lcccc@{}}
\toprule
\multicolumn{2}{l}{\textbf{Method}} & \textbf{MSE}  & \textbf{MAE} & \textbf{MAPE}  \\
\midrule
Linear Regression & &  1.2298 & 0.4757 & 0.2106\\
\multicolumn{2}{l}{ \quad + MQCNN} &  1.3789 & 0.5165 & 0.2486 \\
  \multicolumn{2}{l}{ \quad + GNN} &   0.7633 & 0.4138 & 0.1821\\ 
\multicolumn{2}{l}{Random Forest} & 2.0397 &  0.5051 & 0.2706
 \\ 
\multicolumn{2}{l}{ \quad + MQCNN} & 1.7943 & 0.5639 & 0.3249 \\
\multicolumn{2}{l}{ \quad + GNN} & 1.8419 & 0.5971 & 0.3473 \\
\multicolumn{2}{l}{XGBoost} & 1.7056 & 0.5516 &  0.3091 \\ 
\multicolumn{2}{l}{ \quad + MQCNN} & 1.9021 & 0.5745 & 0.3352 \\
\multicolumn{2}{l}{ \quad + GNN} & 1.7943 & 0.5639 & 0.3249 \\ 
\midrule
\multicolumn{2}{l}{MLP} &  1.2708 & 0.4661 & 0.1945  \\ 
\multicolumn{2}{l}{ \quad + MQCNN} &  1.2520 & 0.4668 & 0.1988  \\
\multicolumn{2}{l}{ \quad + GNN} &  1.2708 & 0.4661 & 0.1945 \\ 
\multicolumn{2}{l}{GRU} & & & \\
\multicolumn{2}{l}{ \quad + MQCNN} & 1.7661 & 0.4780 & 0.2694 \\ 
\multicolumn{2}{l}{ \quad + GNN} & 3.1427 & 0.7104 & 0.2234 \\
\multicolumn{2}{l}{LSTM} & & & \\
\multicolumn{2}{l}{ \quad + MQCNN} & 
3.1850 & 0.6686 & 0.1752 \\
\multicolumn{2}{l}{ \quad + GNN} & 1.2073 & 0.2936 & 0.1694 \\ 
\midrule
\multicolumn{2}{l}{MAML} & 1.0752 & 0.4769 & 0.2345 \\ 
\multicolumn{2}{l}{ \quad + MQCNN} &   3.2979 & 0.5898 & 0.3565  \\
\multicolumn{2}{l}{ \quad + GNN} &  1.0383 & 0.4646 & 0.2201 \\ 
\multicolumn{2}{l}{Reptile} & 4.9341 & 0.8118 & 0.5410 \\ 
\multicolumn{2}{l}{ \quad + MQCNN} &  1.5990 & 0.4457 & 0.1831
  \\
\multicolumn{2}{l}{ \quad + GNN} &  1.2041 & 0.4827 & 0.2287
 \\ 
\multicolumn{2}{l}{MetaSGD} & 1.1941& 0.4837 & 0.2379 \\
\multicolumn{2}{l}{ \quad + GNN} & 1.2667 & 0.4651 & 0.1908 \\
\multicolumn{2}{l}{ \quad + MQCNN} & 1.2111 & 0.4562 & 0.1983 \\ 
\multicolumn{2}{l}{TSA} & 1.7022 & 0.7310 & 0.4936  \\ 
\multicolumn{2}{l}{ \quad + GNN} & 1.5116 & 0.5253 & 0.2585  \\
\multicolumn{2}{l}{ \quad + MQCNN} & 1.1785 & 0.5315 & 0.3084  \\ 
\multicolumn{2}{l}{MUMOMAML} & 0.9449 & 0.3251& 0.2189  \\ 
\multicolumn{2}{l}{ \quad + MCQNN} &  1.0945 & 0.3861 & 0.1920
 \\
\multicolumn{2}{l}{ \quad + GNN} &  0.7577 & 0.3752 & \textbf{0.1553} \\
\multicolumn{2}{l}{F-FOMAML (Ours)} & 0.6552 & 0.3876 & 0.2117 \\ 
\multicolumn{2}{l}{ \quad + MQCNN} &  0.6371 & 0.4134 & 0.2000  \\ 
\multicolumn{2}{l}{ \quad + GNN} &  \textbf{0.6089} & \textbf{0.3713} & 0.2077 \\
\bottomrule
\end{tabular}
\vspace{-2em}
\hspace{1pt} 
\end{table}

\paragraph{Baselines.} Our evaluation encompasses a diverse range of baseline techniques for comparative analysis. This includes traditional regression techniques like Linear Regression, along with ensemble strategies such as Random Forest and the well-regarded XGBoost algorithm \citep{chen2016xgboost}. In the realm of neural-network-based methods, we consider the Multi-Layer Perceptron network, Gated Recurrent Unit (GRU) \citep{chung2014empirical}, Dipole \citep{ma2017dipole}, and LSTNet \citep{lai2018modeling}. Additionally, advanced transfer methods like Model-Agnostic Meta-Learning (MAML) \citep{finn2017model}, Reptile~\citep{Nichol2018OnFM}, Meta-SGD \citep{li2017meta}, TSA~\citep{zhou2020task} and MUMOMAML~\citep{vuorio2019multimodal} are included. Consistency in the feature set is maintained across all baseline models, aligning them with our proposed method, and ensuring a fair comparison.

\paragraph{Model Evaluation and Training.}
With the meta-learning framework in place, we train the base learners on the proxy data and evaluate their performances using evaluation metrics such as mean squared error (MSE), mean absolute error (MAE), and mean absolute percentage error (MAPE). The meta-learner, which could be a neural network~\citep{goodfellow2016deep}, support vector machine~\citep{cortes1995support}, or decision tree~\citep{breiman1984classification}, selects the best base learners and their corresponding hyper-parameters based on the evaluation results.
Next, the selected base learners are fine-tuned on the available historical sales data from the regular sales period, if any, to adapt the model to the peak period. This fine-tuning step allows our model to better capture the unique relationships between features and sales in the target vending machine, leading to more accurate predictions and improved generalization to new tasks.

\subsection{Analysis of Results}

Table~\ref{tab:jd_result} presents a detailed comparison of various machine learning methods applied to real-world e-commerce sales data from JD.com, evaluating their performance through three metrics. The methods encompass traditional regression, ensemble strategies, neural network-based approaches, transfer learning methods, and some advanced meta-learning algorithms, with the inclusion of proxy data features generated by either MQCNN or Graph Neural Networks (GNN).
Linear regression, serving as a baseline, shows moderate performance, which slightly deteriorates when combined with MQCNN but improves with GNN, indicating GNN's effectiveness in feature enhancement. Random Forest and XGBoost, both ensemble methods, exhibit higher errors compared to linear regression, with their performance variably impacted by MQCNN and GNN additions, suggesting a complex interaction between ensemble methods and proxy data features. Among neural network-based methods, the addition of MQCNN generally does not significantly alter performance, whereas GNN integration shows mixed results. Notably, advanced methods like MAML, Reptile, and MetaSGD show varied outcomes, with some combinations leading to increased errors. Particularly, Reptile demonstrates a substantial error reduction when combined with MQCNN, highlighting the potential of integrating advanced algorithms with proxy data feature generation techniques.
The performance of TSA and MUMOMAML, with their respective enhancements, underscores the importance of selecting appropriate proxy data feature generation methods to improve prediction accuracy. MUMOMAML combined with MQCNN achieves the best MAPE score across all methods, emphasizing the strength of multimodal meta-learning techniques when optimized with suitable proxy data features.

The impact of incorporating different proxy data selection methods, particularly the comparison between GNN and other methods like clustering, is profound. The integration of GNN with various machine learning models, including our proposed F-FOMAML, consistently improved performance across metrics (MSE, MAE, MAPE), highlighting the effectiveness of GNN in enhancing the model's ability to predict demand accurately. This improvement is notably apparent in the substantial performance leap observed when F-FOMAML is combined with GNN, which yields the best results. 

\paragraph{Ablation study.} 
To better understand the effect of proxy data, we perform an ablation study by varying the \textit{k} parameter in the \textit{k-shot} proxy data selection and evaluating the performance metrics as the value of \textit{k} changes. As illustrated in Figure~\ref{fig:metrics_eval_k}, we observed that initially increasing \textit{k} leads to a rise in the error metric, suggesting a decline in model performance due to less relevant data. This trend reaches a plateau, after which further increases in \textit{k} result in decreased error, indicating improved performance from a larger proxy data set. These findings highlight a critical threshold where the quantity of proxy data begins to enhance model performance, emphasizing the potential benefits of utilizing larger proxy data sets. Detailed analyses and additional studies on algorithm convergence are provided in the Appendix~\ref{sec:add_exp_JD}.

Our analysis conclusively demonstrates that F-FOMAML, especially when enhanced with Graph Neural Network (GNN) proxy data, outshines traditional regression models, ensemble strategies, neural networks, and other advanced meta-learning algorithms in predicting e-commerce sales on JD.com. This method achieves the lowest MSE and MAE, evidencing its superior ability to capture complex data patterns. Moreover, the integration of GNN as a proxy data selection method significantly boosts F-FOMAML's performance across various metrics, including MSE, MAE, and MAPE, underscoring the pivotal role of advanced proxy data techniques in improving demand prediction models. The comparison with other proxy data methods, such as MQCNN, further illustrates GNN's unique capability in effectively capturing complex data relationships, enhancing F-FOMAML's predictive accuracy. These findings highlight the critical impact of combining GNN with state-of-the-art prediction models, offering insights into the development of more precise demand prediction algorithms.

\section{Conclusion}\label{sec:conclusion}
This paper presents a novel approach to the challenging task of predicting demand during promotional events characterized by special buying behaviors. Traditional sales data often falls short due to the limited availability of historical data for such events. To address this, we framed demand prediction within the graph-augmented meta-learning paradigm. Utilizing the GNN-enhanced F-FOMAML algorithm, which integrates the generalizability of meta-learning with the adaptability of FiLM layers, we developed a robust forecasting model particularly effective in data-sparse scenarios.

Our method is grounded in solid theoretical foundations, demonstrating the algorithm's ability to optimize predictive risk by skillfully managing bias-variance trade-offs. Empirical evaluations highlight our model's superiority over conventional forecasting techniques and underscore its applicability beyond retail, with potential uses in fields such as online banking security and digital marketing. Empirically, F-FOMAML achieves significant improvements, reducing prediction MAE by 26.24\% on the vending machine dataset and 1.04\% on the JD.com dataset, with a notable 10.18\% enhancement over GNN-based benchmarks. 
Further discussion on the strengths, limitations, and future research directions is provided in the Appendix~\ref{sec:discussion}.





\bibliographystyle{ACM-Reference-Format}
\bibliography{ref}

\newpage
\appendix
\section{Appendix}


This appendix offers additional content to complement the core findings of our research. Below is a summary of the sections included:

\begin{itemize}
    \item Proof of Theorem \ref{thm:excess-risk} (Section \ref{sec:proof}): A rigorous, step-by-step proof of Theorem \ref{thm:excess-risk} is presented, solidifying the theoretical underpinnings of our proposed methodology. 
    \item Experimental Details and Supplementary Results (Section \ref{sec:add_exp}): We meticulously outline the datasets employed, the design choices, and the implementation specifics of our experiments, and present additional results that reinforce our conclusions. 
    \item Discussion and Future Directions (Section \ref{sec:discussion}):  We critically discuss the strengths and limitations of our approach,  illuminating potential avenues for future research endeavors.
\end{itemize}

\section{Proof of Theorem~\ref{thm:excess-risk}}\label{sec:proof}
\begin{proof}
 Let us first define an intermediate function:
 \begin{equation*}
h_{\tilde t}^{(im)}=\frac{\sum_{i=1}^n w(\tilde t, t_i)  h_i}{\sum_{i=1}^n w(\tilde t, t_i)}.
 \end{equation*}

 We then define the event $E_n=\{\sum_{i=1}^n w(\tilde t, t_i)>0\}$. Conditioned on the event $E_n$, we have 
 \begin{align*}
\E\big[\big(h^{(im)}_{\tilde t}(f(x))&-\hat h_{\tilde t}(f(x))\big)^2\big]\\
\le&\frac{\sum_{i=1}^n w(\tilde t, t_i)\cdot\E[(\hat h_i(f(x))-h_i(f(x)))^2]}{(\sum_{i=1}^n w(\tilde t, t_i))^2}\\
\le&\frac{\max_i\E[(\hat h_i(f(x))-h_i(f(x)))^2]}{\sum_{i=1}^n w(\tilde t, t_i)}\\
=&O\left(\frac{C(\mathcal H)}{N\sum_{i=1}^n w(\tilde t, t_i)}\right),
 \end{align*}
where the last inequality uses the assumption that 
 $\E[(\hat h_t(f(x))-h_t(f(x)))^2]=O\left(\frac{C(\mathcal H)}{N_d}\right)$ and $N_d\gtrsim N$.

 Moreover, since $\| h_{t_1}- h_{t_2}\|_\infty \le C\cdot \|Z_{t_1}-Z_{t_2}\|\le C h$ when $\|Z_{t_1}-Z_{t_2}\|\le h$, we have that on the event $E_n$, 
\begin{align*}
|h_{\tilde t}^{(im)}-h_t|&=\Big|\frac{\sum_{i=1}^n w(\tilde t, t_i)  (h_i-h_t)}{\sum_{i=1}^n w(\tilde t, t_i)}\Big|\\
&=\Big|\frac{\sum_{i=1}^n \1\{A(\tilde t,t_i)<h\}  (h_i-h_t)}{\sum_{i=1}^n \1\{A(\tilde t,t_i)<h\}}\Big|\\
&\le C h.
\end{align*}
 On the other hand, conditioned on the event $E_n^c$, when the denominator equals to 0, by definition we have $h_t=0$, and therefore
 \begin{align*}
|h_{\tilde t}^{(im)}(f(x))-h_t(f(x))|^2=h_{\tilde t}^2(f(x)).
 \end{align*}
 Consequently, we can write 
\begin{align*}
|h_{\tilde t}^{(im)}(f(x))-h_t(f(x))|^2\le C^2h^2+h_{\tilde t}^2(f(x))\1_{E_n^c}.
\end{align*}
 Therefore,  we have
 \begin{align}\label{eq:h-h}
 \E[\left(\hat h_{\tilde t}-h_{\tilde t}\right)^2]\lesssim &\E\left[\frac{C(\mathcal H)}{N\sum_{i=1}^n w(\tilde t, t_i)}\cdot \1_{E_n}\right]\cr
 &+h^2\cr
 &+\E[h_{\tilde t}^2(f(x))\cdot\1_{E_n^c}].
 \end{align}
 To bound the first term on the right-hand side (RHS), we let
 \begin{equation}\nonumber
 Y=\sum_{i=1}^nw(\tilde t, t_i)= \sum_{i=1}^n\1\{|Z_{\tilde t}-Z_{t_i}\|>h\}.
 \end{equation}
Since $Z_d$ are uniformly distributed on $[0,1]^r$, we have that $Y\sim Binomial(n,q)$ with $q=\mathbb{P}(\|Z-Z_{\tilde t}\|>h)$.
 Using the property of binomial distribution, we have 
 \begin{align*}
\E\left[\frac{\1\{Y>0\}}{Y}\right]\lesssim\frac{1}{nq}\lesssim \frac{1}{n h^r}.
 \end{align*}
 Therefore, the first term on RHS is upper bounded as:
 \begin{align*}
 \E\left[\frac{C(\mathcal H)}{N\sum_{i=1}^n w(\tilde t, t_i)}\cdot \1_{E_n}\right]\lesssim \frac{C(\mathcal H)}{N\cdot nh^r}.
 \end{align*}
 The third term in relation (\ref{eq:h-h}) can be bounded as 
 \begin{align*}
     \E[h_{\tilde t}^2(f(x))\cdot\1_{E_n^c}]&\le \sup h_{\tilde t}^2(f(x)) \E[(1-q)^n]\\
     &\lesssim \sup h_{\tilde t}^2(f(x))\frac{1}{qn} \\
     &\lesssim \frac{1}{nh^r}.
 \end{align*}
 Combining all the pieces, we get 
\begin{equation*}
\E[(\hat h_{\tilde t}-h_{\tilde t})^2]\lesssim h^2+\frac{C(\mathcal H)/N}{nh^r}.
\end{equation*}
Therefore, when $l$ is  Lipschitz with respect to the first argument, we have that 
 \begin{align*}
\E_{(x,y)\sim \tilde t} \left[l(\hat g_{\tilde t}(x;\mathcal D^{tr}, A),y)\right]&-\E_{(x,y)\sim \tilde t} [l(g_{\tilde t}(x),y)]\\
\le&\E[|\hat h_{\tilde t}-h_{\tilde t}|]\\
\le & \sqrt{\E[(\hat h_{\tilde t}-h_{\tilde t})^2]}\\
\lesssim&  h+\sqrt\frac{C(\mathcal H)/N}{nh^r}.
\end{align*}
If we further take $h\asymp (\frac{C(\mathcal H)/N}{n})^{\frac{1}{r+2}}$, we obtain 
\begin{align*}
\E_{(x,y)\sim \tilde t} [l(\hat g_{\tilde t}(x;\mathcal D^{tr}, A),y)]&-\E_{(x,y)\sim \tilde t} [l(g_{\tilde t}(x;\mathcal D^{tr}, A),y)]\\
&\lesssim \left(\frac{C(\mathcal H)/N}{n}\right)^{\frac{1}{r+2}},
\end{align*}
which completes the proof.
\end{proof}

\section{Detailed Experiment Design and Additional Results}\label{sec:add_exp}

\subsection{Dataset Description} \label{sec:dataset}
\paragraph{Vending Machine Company}
In our experiments, we utilize various datasets that contain detailed information about the sales orders and product attributes. 

\textbf{Sales Orders (all those files ended with sale\_order.csv)}: Each dataset covers a specific time range and includes both participating and non-participating products in the experiments. Each record corresponds to a sales order, characterized by the business area, shelf code, order code, product code, user code, purchase date, quantity purchased, sale price, actual payment amount, and product category.

\textbf{Experiment Details (experiment\_detail\_product.csv)}: This dataset includes the products participating in pricing experiments A and B. Each product is characterized by attributes such as the business area, shelf code, scene, scene subdivision, product category, product sub-category, experimental group designation, sale price, adjusted price, cross-price indicator, and an indicator for prices below 95\% of the overall average price.

\textbf{Product Details (product\_detail.csv)}: This dataset contains information about the attributes of each product, including the product code, product category, product type, product sub-category, and an indicator specifying whether it is a common product.

\textbf{Shelf Details (shelf\_detail.csv)}: This dataset provides attributes of each shelf, including the business area, shelf code, an indicator for low-selling devices in the previous month, an indicator for the ability to place high-priced products, old user rate, and shelf grade.

The various data points in these datasets allow for an extensive and comprehensive analysis of the sales patterns and facilitate the learning and evaluation of the proposed model.

\begin{figure}[!ht]
    \centering
    \includegraphics[width=0.8\linewidth]{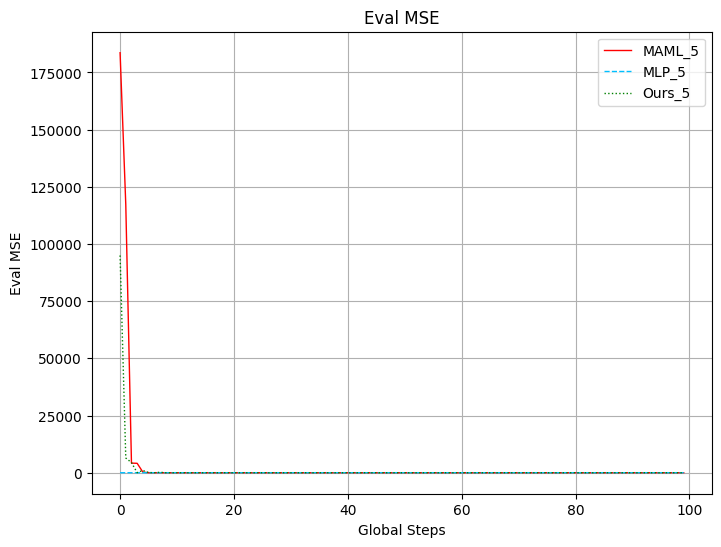}
    \caption{Evaluation Performance (MSE values) throughout training epochs over MAML, MLP and our proposed methods, where $k$ is set to be $5$.}
    \label{fig:eval}
\end{figure}

\begin{table*}[!ht]
\centering
\caption{Detailed feature definition in the Vending Machine dataset.}
\label{tab:sales_order_datasets}
\resizebox{\textwidth}{!}{
\begin{tabular}{|l|l|l|}
\multicolumn{3}{c}{\textbf{(a)} Sales Order Datasets} \\
\hline
\textbf{Feature} & \textbf{Type} & \textbf{Description} \\
\hline
business\_area & String & Area of the business. \\
shelf\_code & String & Unique identifier for the shelf. \\
order\_code & String & Unique identifier for the order. \\
product\_code & String & Unique identifier for the product. \\
user\_code & String & Unique identifier for the user. \\
pay\_date & Date & Purchase date. \\
quantity\_act & Integer & Quantity purchased. \\
sale\_price & Float & Sale price of the product. \\
real\_total\_price & Float & Actual payment amount. \\
product\_type & String & Product category. \\
\hline
\multicolumn{3}{c}{\textbf{(b)} Experiment Detail Dataset}\\
\hline
\textbf{Feature} & \textbf{Type} & \textbf{Description} \\
\hline
business\_area & String & Area of the business. \\
shelf\_code & String & Unique identifier for the shelf. \\
product\_code & String & Unique identifier for the product. \\
mtype & String & Scene. \\
scene & String & Scene subdivision. \\
second\_type\_name & String & Product category. \\
sub\_type\_name & String & Product sub-category. \\
if\_exper & Integer & Indicator for the experimental group (1) or control group (0). \\
sale\_price & Float & Sale price of the product. \\
ab\_price & Float & Adjusted price. \\
cross\_price & Float & Cross-price indicator. \\
lower\_price95 & Integer & Indicator for prices below 95\% of the overall average price. \\
\hline
\multicolumn{3}{c}{\textbf{(c)} Product Detail Dataset} \\
\hline
\textbf{Feature} & \textbf{Type} & \textbf{Description} \\
\hline
product\_code & String & Unique identifier for the product. \\
type\_name & String & Product category. \\
second\_type\_name & String & Product type. \\
sub\_type\_name & String & Product sub-category. \\
is\_common\_product & Integer & Indicator for common products   (1 for yes, 0 for no). \\
\hline
\multicolumn{3}{c}{\textbf{(d)} Shelf Detail Dataset} \\
\hline
\textbf{Feature} & \textbf{Type} & \textbf{Description} \\
\hline
business\_area & String & Area of the business. \\
shelf\_code & String & Unique identifier for the shelf. \\
is\_low\_sale & Integer & Indicator for low-selling devices in the previous month (1 for yes, 0 for no). \\
can\_fill\_high\_price & Integer & Indicator for the ability to place high-priced products. \\
old\_user\_rate & Float & Old user rate. \\
grade & String & Shelf grade. \\
\hline
\end{tabular}
}
\end{table*}

\paragraph{JD.com}

We work with the transactional records from JD.com, which offer a blend of both static and dynamic features related to the product (SKU) and order details for March 2018~\citep{JDdata}.
The \textit{SKU table} contains information about the SKUs that were clicked at least once during March 2018. Each SKU entry has a unique SKU ID and is associated with a seller. For this study, 9,167 SKUs were selected. Each SKU possesses two pivotal attributes, which could, depending on the category, represent product features like SPF for face moisturizers or the number of personalized shaving modes for electric shavers. 

The \textit{Order table} encompasses details about unique customer orders within our designated product category from March 2018. This table elucidates specifics like order quantity, order date and time, SKU type, and the promised delivery time of the order. Additionally, it captures the product pricing and promotional activities, delineating the difference between the original and final unit price, thereby indicating the promotional discounts offered.


For our analysis, we split our tasks into the following partitions: 1) Training set $\mathcal{D}_{train}$, comprising data from regular sales days, this is used for initial model training. 2) Query set $\mathcal{D}_{query}$, which consists of slightly modified versions of tasks from $\mathcal{D}_{train}$ and facilitating the inner-loop adaptation. 3) Testing set $\mathcal{D}_{test}$, that incorporates data from peak sales periods and is earmarked for assessing model performance.






\begin{table*}[!ht] \label{tab:jd-data}
\centering
\caption{Detailed feature description of the JD.com dataset.}
    \resizebox{\textwidth}{!}{
\begin{tabular}{|l|l|l|l|}
\multicolumn{4}{c}{\textbf{(a)} Description of the SKU table} \\
\hline
{\textbf{Field}} & {\textbf{Data type}} & {\textbf{Description}}  & {\textbf{Sample value}}  \\ \hline
{sku\_ID} & {String}  & {Unique identifier of a product}   & {b4822497a5}  \\ 
{type}& {Int}& {1P or 3P SKU}& {1} \\ 
{brand\_ID}      & {String}  & {Brand unique identification code} & {c840ce7809}  \\ 
attribute1      & Int& First key attribute of the category   & 3  \\
attribute2      & Int& Second key attribute of the category  & 60     \\
activate\_date  & String  & The date at which the SKU is first Introduced         & 2018-03-01  \\
deactivate\_date& String  & The date at which the SKU is terminated    & 2018-03-01 \\  \hline
\multicolumn{4}{c}{\textbf{(b)} Description of the users table} \\
\hline
\textbf{Field}      & \textbf{Data type} & \textbf{Description}             & \textbf{Sample value} \\
\hline
user\_ID            & String             & User unique identification code  & 000000f736            \\
user\_level         & Int & User level        & 10     \\
first\_order\_month & String             & First month in which the customer placed an order on JD.com & 2017-07\\
plus & Int & If user is with a PLUS membership& 0      \\
gender              & String             & User gender (estimated)          & F      \\
age  & String             & User age range (estimated)       & 26–35  \\
marital\_status     & String             & User marital status (estimated)  & M      \\
education           & Int & User education level (estimated) & 3      \\
purchase\_power     & Int & User purchase power (estimated)  & 2      \\
city\_level         & Int & City level of user address       & 1   \\  
\hline
\multicolumn{4}{c}{\textbf{(c)} Description of the orders table
} \\
\hline
\textbf{Field} & \textbf{Data type} & \textbf{Description}       & \textbf{Sample value} \\
\hline
order\_ID      & String             & Order unique identification code          & 3b76bfcd3b            \\
user\_ID       & String             & User unique identification code           & 3cde601074            \\
sku\_ID        & String             & SKU unique identification code            & 443fd601f0            \\
order\_date    & String             & Order date (format: yyyy-mm-dd)           & 2018-03-01            \\
order\_time    & String             & Specific time at which the order gets placed             & 2018-03-01 11:10:40.0 \\
quantity       & Int & Number of units ordered    & 1      \\
type           & Int & 1P or 3P orders            & 1      \\
promise        & Int & Expected delivery time (in days)          & 2      \\
original\_unit\_price         & Float              & Original list price        & 99.9   \\
final\_unit\_price            & Float              & Final purchase price       & 53.9   \\
direct\_discount\_per\_unit   & Float              & Discount due to SKU direct discount       & 5.0    \\
quantity\_discount\_per\_unit & Float              & Discount due to purchase quantity         & 41.0   \\
bundle\_discount\_per\_unit   & Float              & Discount due to “bundle promotion”        & 0.0    \\
coupon\_discount\_per\_unit   & Float              & Discount due to customer coupon           & 0.0    \\
gift\_item     & Int & If the SKU is with gift promotion         & 0      \\
dc\_ori        & Int & Distribution center ID where the order is shipped from   & 29     \\
dc\_des        & Int & Destination address where the order is shipped to  & 29    \\
& & (represented by the closest distribution center ID) & \\
\hline
\end{tabular}
}
\end{table*}

\subsection{Implementation Details}\label{sec:exp_impl}

We employ PyTorch for model implementation and Adam for optimization. The learning rate scheduler uses a warmup phase, accounting for 10\% of the total training steps, followed by a linear decay to zero. We set the dropout rate at 0.5 to prevent overfitting during training. 

Each training and test episode corresponds to a single task. For each task, we sample 5 data points (k-shot) for training, validation, and testing. We iterate over 1000 training episodes, 200 validation episodes, and 1000 test episodes.

The hidden dimensions of the product-specific and machine-specific estimators are set to 32. After training for 100 epochs, the model with the lowest validation MSE is saved and used for testing. 

All experiment outputs are saved in a timestamped directory under the project's output path for reproducibility and future reference.

After training, we load the model with the best performance on the validation set to evaluate on the testing set, and report the test scores.

In terms of GNN-based learning tasks,
we aggregate order information for each product to generate the daily quantities sold during the one month of JD.com data. We also include the original and final price sequences as the time series features. Meanwhile, we also obtain each product's static features such as brand and attributes. Our forecasting task is set to predict the sales for the future one day of each product using the past 16 days' information (i.e., $C=16$ in Equation (\ref{eqn:fctask})). 

The graph constructed uses products' brand information such that products under the same brand are connected. 
In this task, the graph contains 9159 nodes that represent all products having at least 1 unit of sale in one month. We use two graph convolutional  (GCN) layers to explore all 2-hop neighbors for each product, together with its own static and dynamic features in the prediction task. 

We train the graph-based model until convergence. The Adam optimizer with default settings is used for minimizing the Mean Absolute Error (MAE) loss with a batch size of 8 to 100 epochs. The final embedding for each product is a numerical vector of length 90 which contains $50, 8, 32$ encoded features representing static, dynamic, and graph-enhanced features respectively.

Our F-FOMAML model is configured with the hyper-parameters in Table~\ref{tab:hyper-parameter}.
 We use two-layer feed-forward neural networks with RELU activation to encode the product and vending machine separately. The hidden size is set to $64$. All models are trained for 100K episodes. Each episode is a regression task. We use the mean-square-error loss. We use Adam as the optimizer with a learning rate of 1e-3. A linear warmup scheduler is used with the first 10\% episodes as the warm-up episodes. The dropout is set to 0.5. We implement our method using PyTorch 1.11 and Python 3.8. The model is trained on a CentOS Linux 7 machine with 128 AMD EPYC 7513 32-Core Processors, 512 GB memory, and eight NVIDIA RTX A6000 GPUs.

\begin{table}[!ht]
    \centering
    \caption{Hyper parameters configuration in our F-FOMAML network.}
    \label{tab:hyper-parameter}
    \begin{tabular}{l|l}
    \toprule
 Dropout rate&  0.5 \\
   Hidden dimension size&  32\\
   Training $k$-shot&  5\\
   Training episodes&  1000\\
   Validation k-shot&  5\\
   Validation episodes&  200\\
   Test $k$-shot&  5\\
   Test episodes&  1000\\
   Epochs&  100\\
   \hline
   Learning rate scheduler&  WarmupLinear \\
   warmup ratio & 0.1 \\
   \hline
   Optimizer&  Adam optimizer \\
   learning rate&  $0.001$\\
   Weight decay&  0\\
   \hline
   Monitor metric &  Mean Squared Error (MSE) \\
   \bottomrule
    \end{tabular}
\end{table}

\subsection{Additional Experiment Results}\label{sec:add_exp_res}
To validate our methodology, we sourced data from two major trading contexts, namely, the vending machine merchandising dataset and the JD.com dataset from a renowned e-commerce platform (i.e., JD.com).

\paragraph{Experiments on Vending machine data}
\textit{Vending Machine Merchandising Dataset.} This dataset is derived from a private vending machine company. The dataset contains sales data from Mar 10, 2022, to April 20, 2022, for 246 products and 1715 vending machines. Each product from a specific vending machine has a base price (last for the first 20 days) and an adjusted price (last for the last 20 days). The goal is to estimate the demand at the adjusted price given the demand at the base price. We use the category information (7 categories in total) for product features, and region (4 regions) and scene (8 scenes) information for vending machine feature. We use the last 10 days as the testing set, and the second to last 10 days as training. The detailed data description can be found in Table~\ref{tab:sales_order_datasets}.

\begin{table}[!t]
\centering
\caption{Experiment results of F-FOMAML using vending machine sales data. Our F-FOMAML obtains the smallest error on the real-world dataset among the competing baselines.}
\label{tab:vending_machine_results}
\begin{tabular}{l|c|ccc}
\toprule
{\bf Method} & {\bf with GNN} & {\bf MSE} & {\bf MAE} & {\bf MAPE}\\
\midrule
Linear Regression &No & 0.6218 & 0.4782 &  0.2900\\
MLP  & No & 0.2811 & 0.2038 & 0.1499\\
MAML & No & 0.2985& 0.2143   &0.1587\\
F-FOMAML (Ours) & No & \textbf{0.2345}& \textbf{0.1532}  & \textbf{0.1206}\\
    \bottomrule
\end{tabular}%
\end{table}
Table~\ref{tab:vending_machine_results} compares our F-FOMAML with Linear Regression and MLP, alongside the meta-learning model MAML, in predicting vending machine sales. While Linear Regression and MLP provide strong baselines, their performance is comparable to MAML, which may suggest MAML's potential overfitting issues in data-limited scenarios. Significantly, F-FOMAML surpasses these models, demonstrating enhanced prediction accuracy.
Our analysis indicates that F-FOMAML improves MAE values by 26.24\% over the existing models on the vending machine dataset, benefiting from domain-specific insights for relation construction. This result underscores F-FOMAML's effectiveness in demand forecasting, highlighting its capacity to optimize prediction accuracy through strategic data utilization. 

In the analysis of the vending machine dataset (Table~\ref{tab:vending_machine_results}), GNNs, MQCNN, and other sequential models were excluded due to the dataset's absence of continuous time-series data, essential for GNNs to produce effective embedding. Therefore, a direct comparison involving GNNs for this dataset is not included.

\paragraph{Additional Ablation studies and analysis on JD.com dataset}\label{sec:add_exp_JD}
\begin{figure}[!t]
\centering
\includegraphics[width=0.8\textwidth]{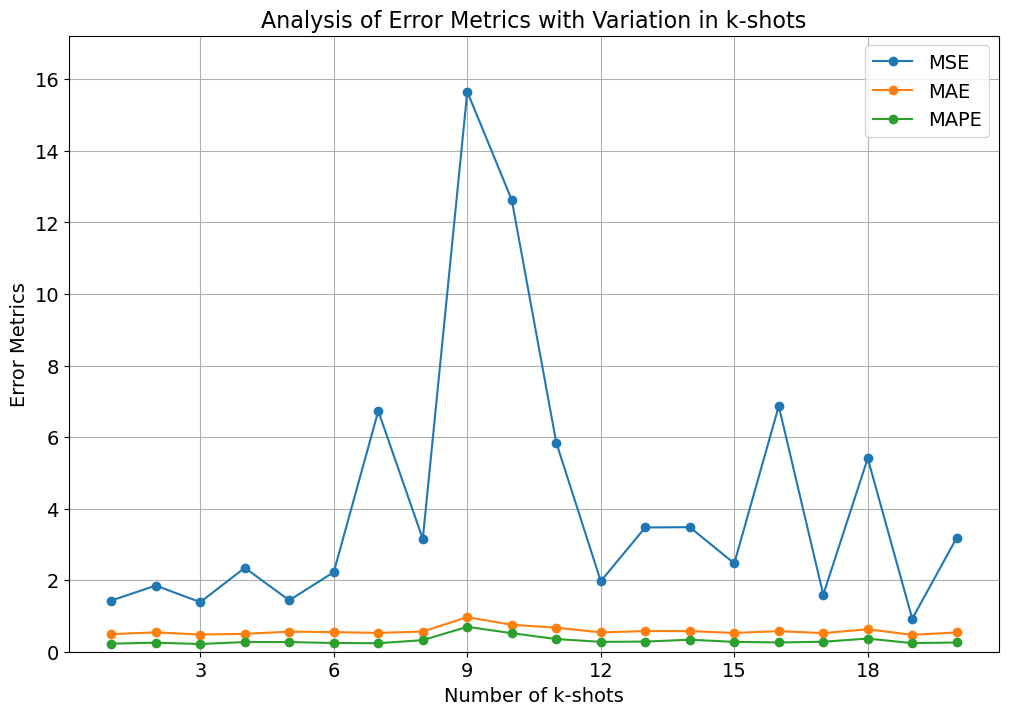}
\caption{Evaluation of metrics versus different values of \textit{k} for the \textit{k-shot} proxy data selection. 
}
\label{fig:metrics_eval_k}
\end{figure}

In our ablation study for proxy data, we aim to unpack the impact of varying volumes of proxy data on our methodology by adjusting the \textit{k} parameter within the \textit{k-shot} proxy data selection framework and examining its influence on our method's error metrics. The outcomes of this investigation are presented in Figure~\ref{fig:metrics_eval_k}, where we meticulously track how changes in the \textit{k} value affect the error metric. Our observations reveal a distinct pattern: as the value of \textit{k} initially increases, there is a corresponding rise in the error metric, suggesting a diminution in model performance possibly due to the incorporation of a larger but less relevant data set. This upward trend in error reaches a plateau, indicating a point of saturation where further increases in \textit{k} do not adversely affect the model to the same extent. Interestingly, beyond this saturation point, the error metric begins to decrease with further increases in \textit{k}, suggesting that the model starts to derive benefits from the expanded pool of proxy data. This indicates the potential value of using larger sets of proxy data, highlighting a critical threshold beyond which the quantity of data begins to outweigh the dilution of relevance, thereby enhancing model performance.

\begin{figure}[!t]
    \centering
    \includegraphics[width=0.8\linewidth]{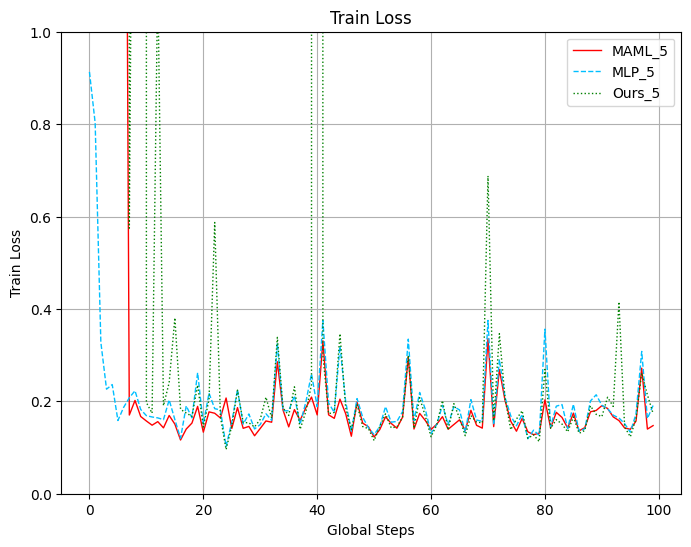}\\
       \vspace{-5pt} 
    \caption{Training performance (MSE values) throughout training epochs over MAML, MLP and our proposed methods, where $k$ is set to be $5$. }
    \label{fig:train-eval}
    \vspace{-8pt}
\end{figure}

Additional details regarding the training performance are presented in Figure~\ref{fig:train-eval}, where our algorithm converges to a lower error than other baselines. 
The fluctuation of our method is due to the feature-based adaptation which results in the stochastic pattern of the convergence behavior.  



    \vspace{-8pt}

\section{Discussion}\label{sec:discussion}
Adaptability and Scalability of F-FOMAML Algorithm: The GNN-enhanced F-FOMAML algorithm is designed for adaptability beyond just demand forecasting, making it suitable for a range of prediction and classification tasks in data-limited scenarios. Its performance on large-scale datasets, like those from JD.com, indicates its scalability. This scalability, alongside the algorithm’s flexibility, suggests potential applicability across diverse industries, showcasing the model's capacity to handle different types of data environments efficiently.Also, our model using the  the first order MAML method. 

As demonstrated in~\citep{Yang2023}, the GNN-based forecaster to generate embeddings can operate in mini-batch fashion and hen can scale to graphs with millions of nodes, which allows the embeddings to be generated based on very large datasets (much larger than the current open source JD.com data). Moreover, our F-FOMAML algorithm is also scalable as it is mathematically similar to Reptile~\citep{Nichol2018OnFM} and also through distributed training which is also demonstrated in the most recent work~\citep{choe2023making}.

Nevertheless, our solution is not devoid of limitations. The success of our methodology is critically tied to the quality and relevance of the proxy data employed. Any deficiencies or omissions in this proxy data could compromise the effectiveness of our approach.

As we gaze into the future, we intend to incorporate meta-learning algorithms specifically designed for regression problems~\citep{Kong2020,Kong2020a}. By doing so, we aim to embed deeper domain-specific knowledge into our model, enhancing its predictive acumen and generalizability. Considering the inherent generality of our methodology, we envision its adaptation to tackle various data-limited scenarios, such as cold-start recommendation or trend forecasting, marking them as potential trajectories for future endeavors.

\end{document}